\documentclass[10pt,twocolumn,letterpaper]{article}

\usepackage{iccv}
\usepackage{times}
\usepackage{epsfig}
\usepackage{graphicx}
\usepackage{amsmath}
\usepackage{amssymb}
\usepackage{adjustbox}
\usepackage{multirow}
\usepackage{enumitem}

\usepackage{amssymb}
\usepackage{color}
\usepackage{xspace}
\usepackage[table,xcdraw]{xcolor}
\usepackage[normalem]{ulem}
\usepackage{bm}
\usepackage{authblk}

\input cgmacros.tex

\definecolor{dkgreen}       {rgb}{0.0,0.5,0.0}
\definecolor{dkcyan}        {rgb}{0.0,0.5,0.5}
\definecolor{dkmagenta}     {rgb}{0.5,0.0,0.5}
\DefAuthor{HS}{dkgreen}
\DefAuthor{MM}{dkcyan}
\DefAuthor{MK}{dkmagenta}
\DefAuthor{AK}{blue}
\DefAuthor{todo}{red}

\newcommand{\net}{\mathcal{G}}

\DeclareMathOperator{\vol}{vol}

\usepackage[pagebackref=true,breaklinks=true,letterpaper=true,colorlinks,bookmarks=false]{hyperref}

\iccvfinalcopy %

\graphicspath{{Image/}}
\DeclareGraphicsExtensions{.png,jpg,.pdf}

\begin{document}

\title{Differentiable Sensor Layouts for End-to-End Learning of\\ Task-Specific Camera Parameters\vspace{-1em}}

\author{Hendrik Sommerhoff\textsuperscript{*}
\qquad Shashank Agnihotri\textsuperscript{**}
\qquad Mohamed Saleh\textsuperscript{*}
\qquad Michael Moeller\textsuperscript{\dag}
\qquad Margret Keuper\textsuperscript{**}
\qquad Andreas Kolb\textsuperscript{*}
\\\textsuperscript{*}Computer Graphics Group, \textsuperscript{**}Visual Computing Group, \textsuperscript{\dag}Computer Vision Group\\University of Siegen
\vspace{-1em}}

\maketitle
\ificcvfinal\thispagestyle{empty}\fi

\begin{figure*}
   \includegraphics[width=\linewidth]{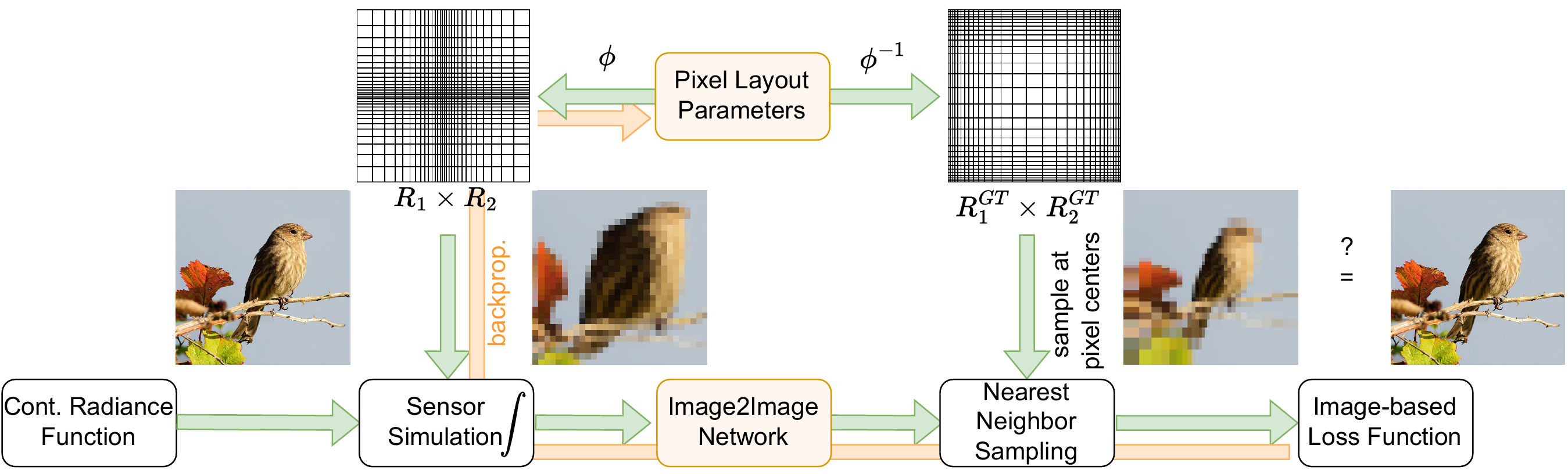}

  \centering

   \caption{The overall pipeline for end-to-end learning of task-specific pixel layouts.}

 \label{fig:teaser}
\end{figure*}

\begin{abstract}
The success of deep learning is frequently described as the ability to train all parameters of a network on a specific application in an end-to-end fashion. 
Yet, several design choices on the camera level, including the pixel layout of the sensor, are considered as pre-defined and fixed, and high resolution, regular pixel layouts are considered to be the most generic ones in computer vision and graphics, treating all regions of an image as equally important.
While several works have considered non-uniform, \eg, hexagonal or foveated, pixel layouts in hardware and image processing, the layout has not been integrated into the end-to-end learning paradigm so far. 
In this work, we present the first truly end-to-end trained imaging pipeline that optimizes the size and distribution of pixels on the imaging sensor jointly with the parameters of a given neural network on a specific task. 
We derive an analytic, differentiable approach for the sensor layout parameterization that allows for task-specific, local varying  pixel resolutions. We present two pixel layout parameterization functions: rectangular and curvilinear grid shapes that retain a regular topology.
We provide a drop-in module that approximates sensor simulation given existing high-resolution images to directly connect our method with existing deep learning models.
We show that network predictions benefit from learnable pixel layouts for two different downstream tasks, classification and semantic segmentation.
\end{abstract}

\section{Introduction}
\label{s:intro}

Deep learning models have achieved impressive performance in a wide range of computer vision tasks, including image classification, object detection, and semantic segmentation.
Traditionally, these pipelines start with an image and consider the parameters of the image formation process, \eg,  pixel layout, pixel gain, and color filters, to be fixed.
Recently, there has been great success with relaxing this assumption and also optimize image signal processing pipeline (ISP) parameters and even lens systems in tandem with neural network parameters in an end-to-end fashion to solve some high-level task \cite{chang18hybrid, chang2019deepoptics, chugunov2021masktof, metzler2020deep, mosleh2020hardware, sun2020learning, tseng2021differentiable}.
Further image formation aspects that have been jointly addressed with a computer vision task are the (multiplexed) color filter layouts for image reconstruction~\cite{chakrabarti2016learning}, and camera calibration, \eg, resolving multi-path interference in time-of-flight imaging~\cite{su2018deep-end-to-end}.

One design decision, that has received relatively little attention in the deep learning community, is the choice of the pixel layout, which refers to the location, size, and shape of pixels on the image sensor. This is surprising since the training of deep learning models is mostly bound by the memory constraints of the GPU, such that it is common practice to artificially decrease the image resolution at the input for network training - without optimizing this process. The vast majority of computer vision tasks are applied to a single universal image model based on a uniform pixel grid.
Since choosing a uniform grid does not make any prior assumption on the distribution of the input data that is to be processed, this layout is the most general.

Several attempts have been made to provide cameras with non-regular pixel layouts. Examples are foveated pixel layouts~\cite{saffih2007foveated} or hexagonal pixels, \eg, to minimize the charge collection time for X-ray detectors~\cite{porro2010expected}. 

Also, looking na\"ively on modern computer vision tasks, we see divergent requirements and different tasks may be better addressed with specifically tailored pixel layouts.
For instance, consider object detection in autonomous driving. A higher pixel density in the center of the image could be beneficial since the spatial frequency of different objects near the horizon is higher compared to the bottom of the image, which is dominated by the street or the car's hood and carries little task-related information.
The optimal layout, in this case, is not immediately apparent and handcrafting it is intractable as it needs to take the downstream network into account, whose output can have a non-trivial dependence on the pixel layout.

To address this problem, we propose a data-driven approach, that optimizes the pixel layout for a given task together with the neural network parameters in an end-to-end fashion.
To the best of our knowledge, this is the first approach for joint optimization of pixel layout and downstream tasks which is realizable by current hardware limitations of sensors.
We validate the effectiveness of our method on classification and semantic segmentation tasks and demonstrate a significant improvement over baselines. Specifically, our contributions are as follows:
\begin{itemize}[itemsep=0pt, parsep=0pt]
    \item We propose a differentiable, physically based sensor simulation framework, that allows end-to-end gradient-based optimization of pixel layouts,
    \item A generic pixel layout parameterization that covers a large class of possible geometries, including rectangular and free-form pixel shapes, while retaining the regular topology, usually required by downstream networks,
    \item A drop-in module, that can approximate the sensor simulation given existing high-resolution images and can thus be easily incorporated into existing deep-learning models, and
    \item We show experimentally that tasks like semantic segmentation in autonomous driving can benefit from non-uniform pixel layouts.
\end{itemize}

\section{Related Work}
\label{sec:related}
First, we discuss prior work that motivates the need for non-uniform pixel layouts.
Then, we consider works that have attempted end-to-end optimization of ISP pipelines for various downstream tasks such as object detection, and semantic segmentation. Some works focus on optimizing sensor parameters for image quality, while others focus on learning effective downsampling.
Lastly, we discuss a different approach for downsampling, \ie, using superpixels and methods that end-to-end optimize superpixel generation and network parameter optimization.

\noindent\textbf{Pixel Layout. } In computer vision tasks like semantic segmentation, most methods employ an ``hour-glass model" like UNet \cite{unet}, PSPNet \cite{zhao2017pyramid}, etc. to be computationally efficient.
Here, the information is first encoded as low-resolution feature maps and then upsampled to image resolution for the downstream task at hand.
However, this approach makes the strong assumption that all regions in the image have equal information, thus equal importance.
In practice, this assumption does not hold as \cite{ultrahighres_iclr2022} shows that a network's performance can benefit from non-regular, learnable downsampling strategies.
Additionally, some early works argue for non-regular pixel layouts due to their ability to improve super-resolution~\cite{ben-ezra2007penrose} or to provide better image representations as such~\cite{kirsch2010precision}.

\noindent\textbf{End-to-end Optimization of the ISP pipeline. }
Some attempts have been made to jointly optimize the ISP pipeline and network parameters \cite{mosleh2020hardware, ultrahighres_iclr2022, object_boundary_high_res, resize_image}.
In their work, \cite{mosleh2020hardware} proposed a ``hardware-in-the-loop" method to jointly optimize the hardware ISP and network parameters as a multi-objective problem using a $0^{th}$-order stochastic solver.
This is orthogonal to our approach as \cite{mosleh2020hardware} optimize over ISP hyperparameters for perceptual image quality, while we optimize to learn the sensor's pixel layout, accounting for the incident pixel radiance.
Some prior work on end-to-end optimization of this problem has been done as well.
\cite{resize_image} use a ``CNN-based resizer" to downsample high-resolution images and then perform classification using deep learning based recognition models and jointly learn weights for the ``resizer" and recognition model.
However, the proposed ``resizer" is extremely complex and not buildable as a sensor, and only performs uniform downsampling.
\cite{object_boundary_high_res} proposed an edge-based downsampling scheme that maps images to a non-uniform pixel layout with a focus on object boundaries, \ie, edges.
They learn to sample with a higher resolution near object boundaries and with lower resolution elsewhere.
This helps in improving the computation cost while retaining some task-essential information.
Nonetheless, many tasks also require information from the object's interior, e.g., cancer cell detection.
To overcome this drawback, \cite{ultrahighres_iclr2022} proposed to learn a deformed sampling density distribution used for downsampling a high-resolution image over a
non-uniform grid based on the network's performance on the downstream task rather than the edge vicinity. 
However, \cite{ultrahighres_iclr2022, object_boundary_high_res} change the pixel layout for each image individually which might lead to poor generalization, while we learn pixel layout over an entire dataset. Moreover, our physically-based sensor simulation considers the full radiance, not just a single sampling position.
At the same time, our drop-in module can take existing high-resolution images to approximate the incident radiance, allowing to optimize any network for a limited resolution budget or to reduce the network's size and training time while keeping performance as high as possible.
In prior work \cite{riad2022learning}, similar approaches have been taken by decreasing the regular pixel resolution of the input images to reduce the network size until the performance falls below a given lower bound.
Our approach can thus be seen as a generalization of \cite{riad2022learning}, integrating non-regular sampling schemes like \cite{ultrahighres_iclr2022}, while  considering the total intensity of the input image (as an approximation of the incident radiance to a sensor).

\noindent\textbf{Superpixel-based Methods.}
Apart from learning sensor parameters for effective downsampling, other approaches  learn pixel grouping into larger superpixels and end-to-end optimizing to learn this grouping and network parameters \cite{superpix,Fey_2018_CVPR,https://doi.org/10.48550/arxiv.2002.05544}. Yet, all such approaches create \textit{image specific} superpixels and are therefore not suitable for learning a \textit{fixed} image sensor layout.
Moreover, as inspired by \cite{chang18hybrid, chang2019deepoptics, chugunov2021masktof, metzler2020deep, mosleh2020hardware, sun2020learning, tseng2021differentiable} our objective rather is to optimize sensor parameters (or image downsampling) in a photometrically correct way, \ie, considering the continuous acquisition of radiance at the pixel level.

\section{Differentiable Sensor Simulation}
As opposed to usual deep learning techniques, which pick a task-specific network architecture $\net$ to make predictions $\net(I, \nu)$ on input images $I$ for parameters $\nu$ of the network, we propose to consider the images $I$ as functions $I(\theta,L_i)$ of the incoming radiance $L_i$ as well as a parameter vector $\theta$ that describes the diffeomorphic deformation of the pixel layout the image $I$ is recorded with. This naturally requires modeling a spatially continuous sensor simulation process that determines the value of each pixel for a given deformation described by $\theta$. As soon as the sensor model, \ie, the dependence of the recorded image $I$ as a function of the deformation parameters $\theta$ and the radiance $L_i$ is determined, we propose to \textit{jointly} train our system for network as well as sensor system parameters by optimizing
\begin{align}
\label{eq:training}
    \min_{\theta, \nu} ~\mathbb{E}_{(L_i,y)}(\mathcal{L}(\net(I(\theta,L_i), \nu),y)), 
\end{align}
where $\mathcal{L}$ is a suitable loss function to compare the network's prediction to the desired prediction $y$.  

Our overall approach for representing $I$ as a function $I(\theta, L_i)$ is depicted in Fig.~\ref{fig:teaser}. Assuming a \emph{continuous radiance function} (or a high resolution image) $L_i$ as input, we perform a \emph{sensor simulation} (see Sec.~\ref{s:method}) to capture the radiance hitting the individual pixels defined by our \emph{pixel layout parameters} (see Sec.~\ref{sec:pixel_layouts}). 
Our pixel layout is a continuous, bijective function $\phi$ applied to a regular pixel grid that retains the pixel topology. Thus, the resulting, usually distorted pixel layout can be fed to any network that accepts images as input.
In case of a downstream image-to-image network that uses an image-based loss function, we apply a \textit{back-warping}, \ie an interpolation using $\phi^{-1}$, to resample the network's output to the original image resolution.
In case of a classification network, no back-warping is required.
As our pixel layout model is differentiable, the optimal pixel layout can be learned jointly with the network (see Sec.~\ref{s:end-to-end-opt}). 

\subsection{Measurement Equation}
\label{s:method}
Here we present the image formation process, specifically the measurement on the sensor plane.
We denote pixels with the multi-index $k = (k_1, k_2)$, e.g., as $I_k$ for the RGB value of pixel $(k_1, k_2)$.
An individual sensor pixel measures the energy $E_k$ of incoming radiance $L_i$, integrated over all possible directions $\omega\in\Omega\subseteq\mathbb{R}^2$, locations $p\in A_k\subseteq\mathbb{R}^2$ inside the pixel area $A_k$, time $t\in [t_0, t_1]$ inside the exposure window and measurable wavelengths $\lambda \in [\lambda_0, \lambda_1]$:
\begin{equation}
    E_k = \int_{\lambda_0}^{\lambda_1 }\int_{t_0}^{t_1}\int_\Omega\int_A W(p, \omega, t, \lambda)L_i(p, \omega, t, \lambda)dp d\omega dtd\lambda.
\end{equation}
Here, $W$ is a weighting term that models the varying responsivity of the sensor with respect to the parameters, like the location inside a pixel or sensitivity to specific wavelengths.
The measured energy is then further processed in an image signal processing pipeline (ISP) to compute an RGB value $I_k\in\mathbb{R}^3$.

In our approach, we make the following simplifying assumptions.
First, we assume a static scene that is captured through a pinhole camera. Further, we interpret the radiance function to be independent of the wavelength and instead output RGB values. This allows us to directly compute $I_k$ as the mean RGB value over the pixel area:
\begin{equation}
    I_k = \frac{1}{\vol({A_k})}\int_{A_k}W(p)L_i(p)dp
\end{equation}
Note, that in Computer Graphics literature, it is often (implicitly) assumed that each pixel has unit area, thus removing the factor in front of the integral.
Since we want to model pixels of different sizes, we have to keep the normalization factor.
In the general case, this normalization is part of the ISP that maps energies to RGB values.

\subsection{Parameterizing Pixel Layouts}
\label{sec:pixel_layouts}
 In the following, we introduce a general framework to parameterize the pixel layouts.
We set the full sensor area to be the square region $\mathcal{S} = [-1, 1]^2$. As such, each pixel has its own domain $A_k \subset \mathcal{S}$ and weighting function $W_k$. We assume pixels to be disjoint, except for their boundaries $\partial A_k$, and their union to cover the whole sensor area, \ie
\begin{equation}
    \mathcal{S} = \bigcup A_k
\end{equation}
We denote pixels from the standard, uniform layout as $U_k$.
In this case, the pixel boundaries match the parallel grid lines of the sensor, \ie
\begin{equation}
\begin{split}
    U_k &= \left[\frac{2k_1 - R_1}{R_1}, \frac{2(k_1+1) - R_1}{R_1}\right]\\
    &\times \left[\frac{2k_2 - R_2}{R_2}, \frac{2(k_2 + 1) - R_2}{R_2}\right], 
\end{split}
\end{equation}
where $R_1, R_2$ is the sensor resolution. Note, that we only modify the pixels' geometry and keep their topology intact

We now define a class of pixel layouts to be a parameterized deformation function $\phi: \mathcal{S} \times \mathcal{D}\rightarrow \mathcal{S}$, where $\mathcal{D} \subset \mathbb{R}^d$ is the set of possible parameters. We require the function $\phi(\cdot, \theta)$ to be bijective and bi-Lipschitz (such that the function and its inverse are also differentiable almost everywhere) for fixed $\theta\in\mathcal{D}$.
For every $\theta$ the function $\phi$ maps the sensor to itself.
We define the pixels under this layout as $A_k(\theta) = \phi(U_k, \theta)$, i.e., the image of the uniform pixel area.

The bijectivity constraint assures that the deformed pixels still do not overlap and the overall pixel number and topology is retained. Furthermore, since $\phi$ is Lipschitz, neighboring pixels are mapped to neighboring pixels again
and boundaries get mapped to boundaries, i.e., $\partial A_k(\theta) = \phi(\partial U_k, \theta)$.

The pixel color  is computed by change of variables
\begin{align}
        I_k(\theta) &= \frac{1}{\vol(A_k(\theta))}\int_{A_k(\theta)} W_k(p, \theta) L_i(p) dp
        \label{eq:change-of-variables}\\\notag
        &= \frac{\displaystyle \int_{U_k} W_k(\phi(u, \theta), \theta) L_i (\phi (u, \theta)) | \det J_\phi (u, \theta)| du}{\int_{U_k} |\det J_\phi (u, \theta)| du}
\end{align}
Here $J_\phi$ is the $2\times2$ Jacobian matrix of $\phi$ with respect to the spatial arguments.
Since the weighting function may depend on the pixel layout, we explicitly added $\theta$ to its arguments.

Eq.~\eqref{eq:change-of-variables} is the forward pass of the sensor's image formation process and can be computed using numerical integration methods. Note, that at this point, we make no assumption on how to sample the incoming radiance $L_i$ and treat it like a black-box.
For instance, it could be computed through recursive path tracing as part of a rendering algorithm.

\begin{figure}
    \centering
    \includegraphics[width=\linewidth]{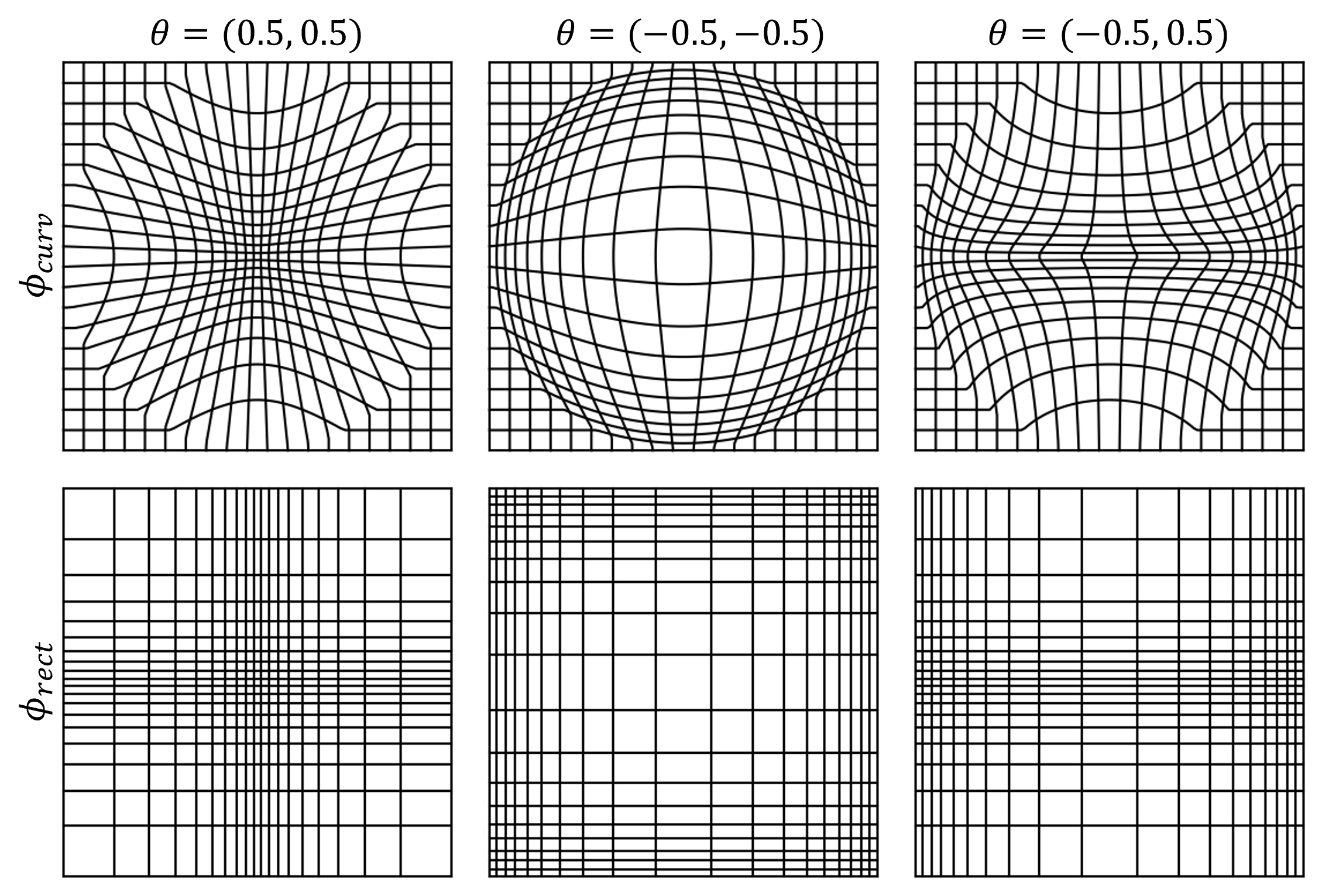}
    \caption{The curvilinear (top row) and rectangular (bottom row) pixel layouts used in this paper and their behavior under different parameters on a $20\times 20$ sensor.}
    \label{fig:deforms}
\end{figure}

There are many possible choices for $\phi$, with varying degrees of freedom with respect to $\theta$.
In this paper, we explore two simple deformation functions that depend on only two parameters each.
Still, in our experiments, we found that such simple pixel layouts already lead to significant improvements.
First, we consider the function
\begin{equation}
    \phi_{curv} (p, \theta) = \begin{cases}
    \begin{pmatrix}
        \frac{p_1(\theta_1-1)}{2\theta_1||p||_2 - \theta_1 - 1}\\
        \frac{p_2(\theta_2-1)}{2\theta_2||p||_2 - \theta_2 - 1}
    \end{pmatrix}& \text{if } ||p||_2 < 1\\
    \begin{pmatrix}
    p_1 \\ p_2
    \end{pmatrix}
    & \text{otherwise}
    \end{cases}
    \label{eq:phi_curv}
\end{equation}
for $p = (p_1, p_2)\in\mathcal{S}$ and $\theta = (\theta_1, \theta_2) \in (-1, 1)^2$.
This function is differentiable with respect to $p$ almost everywhere, except for points on the unit circle, where it has a kink.
The parameters control the vertical and horizontal deformation strengths independently and, dependent on their sign, move more pixels to the edges or center of the unit circle (see top row of Fig.~\ref{fig:deforms}).
This results in a curvilinear grid, which means the pixels are not rectangular anymore.
Since rectangular pixels have advantages, such as easier manufacturing, we also propose $\phi_{rect}$ by making a small modification to Eq.~\ref{eq:phi_curv}:
\begin{equation}
    \phi_{rect}(p, \theta) = \begin{pmatrix}
        \phi_{rect, 1}(p_1, \theta_1)\\
        \phi_{rect, 2}(p_2, \theta_2)
    \end{pmatrix}
\end{equation}
where
\begin{equation}
    \phi_{rect, j}(p_j, \theta_j) = \begin{cases}
        \frac{p_j(\theta_j - 1)}{2\theta_j |p_j| - \theta_j - 1} & \text{if } |p_j| < 1\\
        p_j & \text{otherwise}
    \end{cases}
\end{equation}
for $j\in\{1, 2\}$, \ie., we replace the norm in Eq.~\eqref{eq:phi_curv} with the absolute value of each individual component. The bottom row of Fig.~\ref{fig:deforms} depicts a rectangular pixel layout generated using $\phi_{rect}$.

\subsection{End-to-End Optimization}
\label{s:end-to-end-opt}
\begin{figure}
    \centering
    \includegraphics[width=\linewidth]{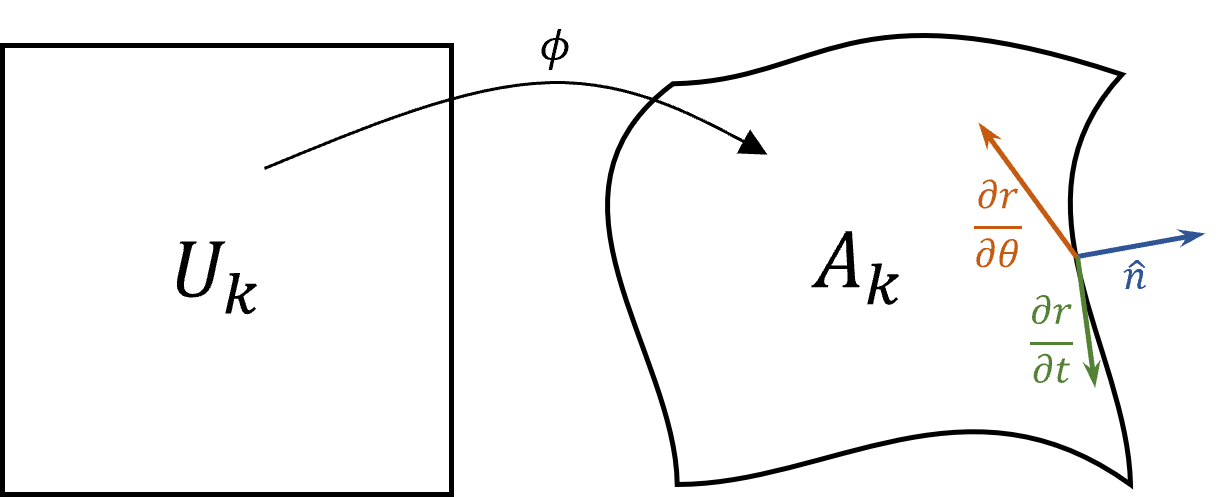}
    \caption{Applying $\phi$ to a uniform pixel $U_k$ yields the deformed pixel $A_k$. The flux integral for gradient computation requires quantities on each boundary point $r\in\partial A_k$: The tangent $\frac{\partial r}{\partial t}$, the outward-pointing normal $\hat{n}$, and the change of the boundary point with respect to the pixel layout parameterization $\frac{\partial r}{\partial \theta}$.}
    \label{fig:boundary_vectors}
\end{figure}
Since the goal is to optimize the pixel layout for some downstream task, we will now explain how to incorporate our pixel layout parameterization in a gradient-based optimization scheme.

After the pixel-outputs of the sensor simulation $I_k(\theta)$ (see Eq.~\eqref{eq:change-of-variables}) are processed by a neural network $\mathcal{G}$, a task dependent loss function $\mathcal{L}$ is computed.
We need to compute the gradient of this loss function with respect to the sensor parameter $\theta$. Applying the chain rule yields
\begin{equation}
    \frac{\partial \mathcal{L}}{\partial \theta_j} = \sum_k \frac{\partial \mathcal{L}}{\partial I_k} \frac{\partial I_k}{\partial \theta_j},
\end{equation}
where $\frac{\partial \mathcal{L}}{\partial I_k}$ can be computed via standard backpropagation as implemented in all common deep learning frameworks. 

In the following, we give an analytic expression for the second term, \ie the derivative of a pixel color with respect to the sensor layout. For brevity, we will drop the $j$ subscript so that $\theta$ always refers to a scalar value. However, the calculations can be easily vectorized.

Starting from Eq.~\eqref{eq:change-of-variables}, applying the quotient rule yields
\begin{align}
    \frac{\partial I_k}{\partial \theta} &= \frac{\partial}{\partial \theta}\left(\frac{1}{\vol(A(\theta))}\int_{A_k(\theta)} W_k(p, \theta) L_i(p) dp\right)\\
    &=: \frac{\partial}{\partial \theta} \frac{f(\theta)}{g(\theta)}\\
    &= \frac{\frac{\partial f}{\partial \theta}(\theta) g(\theta) - f(\theta)\frac{\partial g}{\partial \theta}(\theta)}{g(\theta)^2}.
\end{align}
We explain the derivative of $f$, \ie,  
\begin{equation}
    \frac{\partial f}{\partial\theta} = \frac{\partial}{\partial \theta}\int_{A_k(\theta)} W_k(p,\theta)L_i(p)dp.
\end{equation}
in detail, the derivative of the volume can be computed analogously. 
Since the integration domain itself depends on $\theta$, we can apply Reynold's transport theorem \cite{flanders1973differentiation}:
\begin{equation}
\begin{split}
    \frac{\partial f}{\partial \theta} &= 
    \int_{A_k(\theta)}\frac{\partial }{\partial\theta}W_k(p, \theta) L_i(p)dp\\
    &+ \oint_{\partial A_k(\theta)}W_k(r,\theta)L_i(r) \langle \frac{\partial r}{\partial \theta}, \hat{n}(r, \theta) \rangle ds\\
    &=: Q_{int}(\theta) + Q_{bound}(\theta)
\end{split}
\label{eq:pixel-derivative}
\end{equation}
The first integral can be easily computed by a change of variables like Eq.~\eqref{eq:change-of-variables}.
The second integral is the flux integral across  $\partial A_k$, where $r: [0, 1] \rightarrow \partial A_k(\theta)$ is an arbitrary piece-wise smooth parameterization of the boundary, $\frac{\partial r}{\partial \theta}$ is the local change of the boundary point with respect to $\theta$ and $\hat{n}$ is the corresponding outward pointing unit normal vector at each boundary point. Since the boundary of $A_k$ is exactly the image of the boundary of $U_k$ under $\phi$, a parameterization can be given by
\begin{equation}
    r(t, \theta) = \phi(\gamma(t), \theta)
\end{equation}
where $\gamma$ is a parameterization of the square boundary of $U_k$. Furthermore, tangent vectors of $r$ are the pushforward of tangent vectors of $\gamma$ by $\phi$, i.e., $\dot{r}(t, \theta) = J_\phi \dot{\gamma} (t, \theta)$. Thus, we can compute the line element $ds = || \dot{r}(t, \theta) ||dt$ and the normal (up to orientation) as
\begin{equation}
    \hat{n}(t, \theta) =
    \begin{pmatrix}
        \dot{r}_2(t, \theta)\\
        -\dot{r}_1(t, \theta)
    \end{pmatrix} / || \dot{r}(t, \theta) ||_2
\end{equation}
The important quantities of the boundary integral are visualized in Fig.~\ref{fig:boundary_vectors}.
With this, we have all ingredients to compute the derivative using Eq.~\eqref{eq:pixel-derivative}. 
The interior integral $Q_{int}$ looks very similar to the pixel color formula in Eq.~\eqref{eq:change-of-variables} and can be efficiently computed by reusing radiance samples from the forward pass. The boundary integral $Q_{bound}$ only requires sampling the border of the pixel, which generally requires less samples.
It is also worth noting that neighboring pixels always share a boundary whose normals have opposing orientation.
This means that the absolute value of the line integral at the boundaries of neighboring pixels is the same and only one of the two needs to be computed.

Note, that applying the divergence theorem on $Q_{bound}$ would transform it into another interior integral.
However, in that case we would also need to compute the gradients of $L_i$ with respect to the location on the sensor plane.
Using the formulation above, we do not need to make any assumption on the availability of gradients of $L_i$.

\section{Implementation Details}
\label{sec:implementation}
We implement both, the forward and backward pass, using numerical integration. We employ stratified Monte-Carlo integration to reduce possible aliasing artifacts that could emerge from quadrature-based integration schemes. %
We make no assumptions on how $L_i$ is calculated.
In general, our method can be built on top of any existing rendering algorithm, as long as it exposes a way to sample the radiance at given points on the image plane.

The main goal of our method is to enhance a downstream deep learning task, which generally requires a substantial amount of training data.
For many tasks, large and widely-used datasets are available, whose effectiveness has been proven over time.
We therefore added the alternative option to approximate the radiance function $L_i$ with high-resolution real images, which we transform into a coarser image using our differentiable sensor simulator.
We use bilinear interpolation to sample the radiance at arbitrary positions on the sensor plane. Furthermore, we set $W_k$ to be constant, \ie all points on a pixel have the same sensitivity.

We implement our differentiable sensor as a custom PyTorch layer, which makes it easy to integrate into already existing projects.
The layer takes a high-resolution image as input and outputs an image with the resolution of the simulated sensor.
We use $\tanh$ to restrict the parameters to the allowed range of $(-1, 1)^2$. We will make our implementation of the differentiable sensor layer as well as code for reproducing the experiments shown in this paper publicly available upon acceptance. 

Even though we simulate non-uniform layouts, we interpret the output of the sensor layer as a uniform grid, \ie, the output is just an image tensor in $\mathbb{R}^{3\times R_1 \times R_2}$ without information about the size and relative distance of neighboring pixels.
Directly visualizing this output makes the image look deformed (see Fig.~\ref{fig:teaser}).
For instance, layouts with more pixels in the image center would result in enlarged objects in the image center.
We feed this deformed image directly into the downstream network.
Even though this seemingly violates the implicit assumption of convolutional layers that all pixels have the same distance to their neighbors, our experiments show that we still achieve significant improvements over baselines.
Exploring more tailored solutions accounting for relative pixel positions, like continuous CNNs, should be addressed in future research.

Special care has to be taken if the downstream network is solving an image-to-image task.
In that case, there is usually a one-to-one correspondence between input pixels and output pixels, like a dense map of class probabilities in the case of semantic segmentation.
This means that a change in the pixel layout of the input image has an immediate effect on the output image, which has to be taken into account when computing the loss.
This will, in general, be some form of distance to a ground truth image given in uniform pixels.
We therefore modify the loss function to properly account for pixel area. To this end, we warp a uniform grid with the ground truth image resolution $R_1^\text{gt}\times R_2^\text{gt}$ via the inverse pixel layout deformation $\phi^{-1}(p, \theta)$ and sample the (deformed) output at these locations with nearest-neighbor interpolation.
The result is an un-deformed image with constant regions that match the pixel layout defined by $\phi$.
We now take the loss between this resampled output and the ground truth image as usual.
Note, that we only apply this resampling to the \emph{output} of the downstream task while the network operates on the deformed image and can thus take advantage of the non-uniform layout. 

These considerations only apply to image-to-image tasks.
If the output of a task does not depend on positions on the image plane (for instance classification tasks), the loss function can be applied as-is without modifications.

\section{Experiments}
\label{s:eval}

\paragraph{MNIST classification}
\begin{figure}
    \centering
    \begin{tabular}{cc}
    $\theta = (0, 0)$ & $\theta=(0.56, 0.38)$\\
    \includegraphics[width=0.4\linewidth]{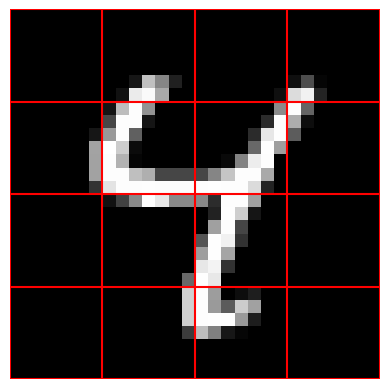}
    &\includegraphics[width=0.4\linewidth]{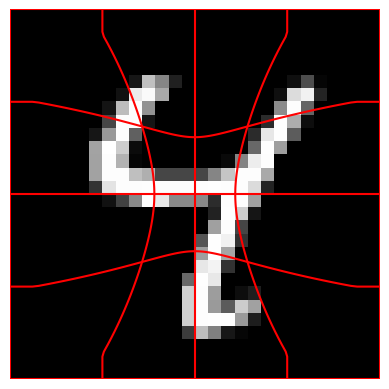}
    \\
    \includegraphics[width=0.4\linewidth]{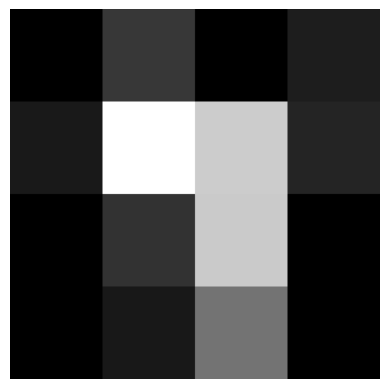}
    &\includegraphics[width=0.4\linewidth]{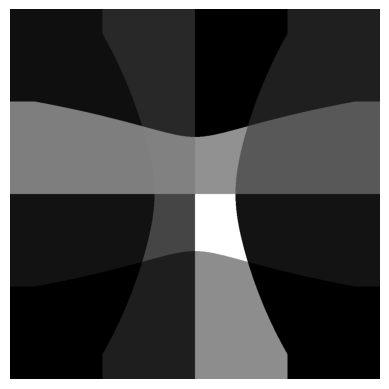}
    \\
    (a) Uniform & (b) Learned
    \end{tabular}
    \caption{Learned sensor layout for $4\times4$ MNIST classification. Top row: pixel layouts overlayed over the original $28\times 28$ image. Bottom row: Simulated sensor output. The network fine-tuned on the uniform layout wrongly classifies (a) as a 7 while the end-to-end optimized network classifies (b) as a 4.}
    \label{fig:mnist_layout}
\end{figure}

\begin{table}[]
    \centering
    \begin{tabular}{c|c}
         Layout & Accuracy\\
         \hline
         Uniform & 83.63\% \\
         Learned (Ours) & \textbf{90.90}\%
    \end{tabular}
    \caption{Test set accuracy of $4\times4$ MNIST \cite{deng2012mnist} classification. Even with a resolution of only $4\times 4$ pixels, we can reach a classification accuracy of more than 90\%.}
    \label{tab:mnist_result}
\end{table}
To illustrate the principle of our end-to-end sensor layout optimization, we start with a toy example and optimize the layout for hand-written digit recognition on MNIST~\cite{deng2012mnist} with a sensor size of only $4\times4$ instead of the original $28\times 28$ pixels.
The digits in MNIST are always centered, so the hypothesis is that an optimized layout puts smaller pixels in the middle in order to capture the higher information density there.
We apply the curvilinear layout $\phi_{curv}$ whose parameters we initialize with $0$, \ie, we start the training with a uniform pixel layout.

The sensor layer output is fed into a small CNN consisting of two convolutional layers with 32 and 64 channels, followed by a max pooling layer and two fully connected layers with hidden dimension 128.
We train the whole pipeline for 14 epochs with Adam~\cite{kingma2014adam} and an initial learning rate of 0.01.
As baseline, we compare against exactly the same pipeline and training scheme, but with frozen uniform sensor parameters.

The test set accuracy in Tab.~\ref{tab:mnist_result} demonstrates a clear advantage of our learned layout, which can be seen in Fig.~\ref{fig:mnist_layout}.
The layout confirms our hypothesis that smaller pixels in the center are advantageous for classification on this specific dataset.
Note that the visualization of the learned layout in Fig.~\ref{fig:mnist_layout} is not what the network receives as input.
The CNN has no notion of pixel shape and sensor layout, and only ``sees'' a $4\times 4$ grid of pixel values.
\vspace{0.1cm}

\noindent\textbf{Semantic segmentation}
\begin{table}[]
    \centering
    \begin{tabular}{c|c|c|c}
        
         Resolution & Method & mIoU (\%)  & Acc. (\%) \\
         \hline
         \multirow{3}{*}{$129\times 65$}&PSPNet \cite{zhao2017pyramid} & 37.11 & 87.17\\
         & PSPNet + $\phi_{curv}$ & 37.56 & 87.33\\
         & PSPNet + $\phi_{rect}$ & \textbf{39.44} & \textbf{87.90}\\
         \hline
         \multirow{3}{*}{$257\times 129$}&
         PSPNet & 50.76& 90.86 \\
         &PSPNet + $\phi_{curv}$ & 52.39& 91.19 \\
         &PSPNet + $\phi_{rect}$ & \textbf{54.08}& \textbf{91.44} \\
         \hline
         \multirow{2}{*}{$513\times 257$}&PSPNet & 63.98 & 93.73\\
         &PSPNet + $\phi_{curv}$ & 64.65& 93.73 \\
         & PSPNet + $\phi_{rect}$ & \textbf{64.74} & \textbf{93.75}
    \end{tabular}
    \caption{Semantic segmentation results with different sensor resolutions on the Cityscapes dataset~\cite{Cordts2016Cityscapes}. For all resolutions, the learned rectangular layout performs best.}
    \label{tab:result_cityscapes}
\end{table}
\begin{figure*}
\centering
\includegraphics[width=\linewidth]{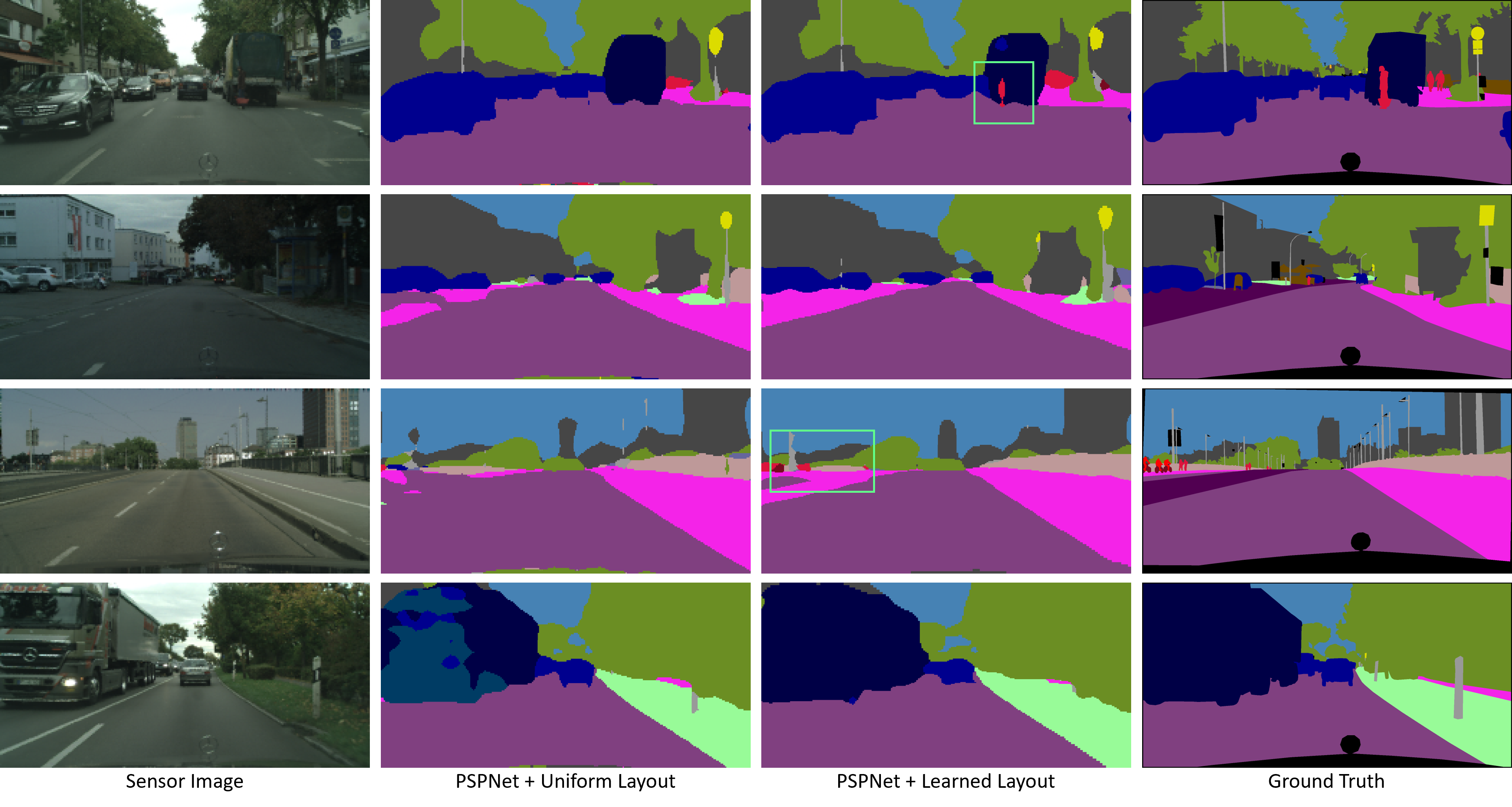}

\vspace{-0.1cm}
\caption{Example segmentations from the Cityscapes~\cite{Cordts2016Cityscapes} test set with and without learned sensor layouts at $257\times 129$ resolution. The used deformation is $\phi_{rect}$. To accommodate the different pixel shapes, both variants were upsampled using nearest neighbor interpolation to the ground truth resolution ($2048\times1024$), as described in Sec.~\ref{sec:implementation}. The learned pixel layout achieves more accurate segmentations with objects detected that were otherwise missed by using the uniform layout, especially in the vertical center of the image.}
\label{fig:cityscapes_qualitative}
\end{figure*}
\begin{figure}
    \centering
    \footnotesize
    \begin{tabular}{@{}c@{}c@{}}
         $\theta = (0.06, 0.33)$ & $\theta = (-0.14, 0.25)$\\
         \includegraphics[width=0.5\linewidth]{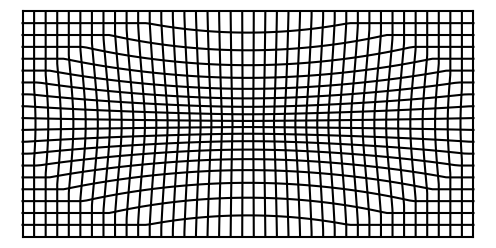} &
         \includegraphics[width=0.5\linewidth]{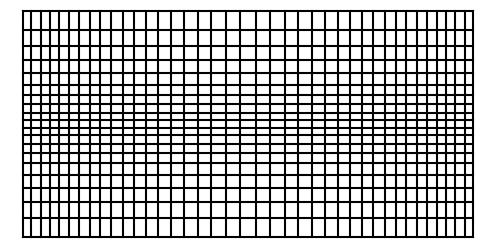}\\
         (a) Curvilinear & (b) Rectangular
    \end{tabular}
    \caption{Learned curvilinear and rectangular pixel layouts for $257\times 129$ segmentation on Cityscapes (less pixels are shown for visualization purposes). Vertically, the sensor learned to put more pixels in the center. Further, the rectangular layout has a slightly higher pixel density towards the left and right of the sensor.
    \vspace{-1em}
    }
    \label{fig:layouts_cityscapes}
\end{figure}
For a relevant practical example application, we evaluate the learned pixel layouts under the task of semantic segmentation of urban street scenes on the Cityscapes dataset~\cite{Cordts2016Cityscapes}.
The dataset contains 5000 high-resolution, densely annotated frames with 19 classes.

Unlike in more general segmentation tasks, street view images are taken in environments that share many similarities between scenes and the distribution of object classes over the image plane is highly non-uniform.
For instance, often large parts of the image are occupied by empty streets or the sky, while others have a higher density of different object classes.
Thus, the segmentation task could benefit from distributing pixels to areas of higher variance.

As a baseline, we choose PSPNet~\cite{zhao2017pyramid} with a ResNet50 backbone~\cite{resnet}.
In general, segmentation networks are trained and evaluated on patches of the high-resolution input images, instead of downsampling them.
Since our objective is to optimize the pixel layout over the whole sensor plane, we instead feed the whole input to the network directly.
This also means that we cannot employ all data augmentation schemes from~\cite{zhao2017pyramid}, namely random rotation.

Again, we train the baseline with a fixed uniform pixel layout and compare it to the same network architecture with a learnable layout.
To assess if curvilinear or rectangular layouts have an advantage over the other, we train variants with both.
Additionally, we evaluate the behavior under different sensor resolutions.

As explained in Sec.~\ref{sec:implementation} we modify the loss function for image-to-image tasks using a resampling procedure on the network outputs, upscaling them to the resolution of the ground truth segmentations ($2048\times 1024$).
For a fair comparison, we do the same resampling to the uniform baseline. 

We employ the exact same training scheme for all experiments that follows the protocol (including default parameters) from \cite{zhao2017pyramid}, except that we use a batch size of 4 instead of 16.
We initialize all sensor parameters with 0 (uniform layout) and freeze training of sensor parameters for the first 10 epochs to speed up convergence to reasonable network weights.
In total, we train each network for 200 epochs.

The learned layouts in Fig.~\ref{fig:layouts_cityscapes} display a higher pixel density towards the horizon line.
As expected, fewer pixels are needed to detect the street and the sky.
Perspective also dictates that objects far away from the camera gather at the horizon and are smaller, thus requiring a finer pixel raster to distinguish them.
Interestingly, the sensor with the rectangular layout also learned to put slightly more pixels towards the left and right edges.
This might be due to the fact that there is often a larger variety of small objects on the sidewalks (people, poles, fences, vegetation) as opposed to the street itself, where there are mostly cars.
The learned curvilinear layout does not show this horizontal non-uniformity. An explanation for this could be that the deformation here is restricted to a circular region, which means that the region at the sensor boundary can not be samples more densely. %

Even though the learned layouts are not as extreme as for the MNIST experiment, the quantitative results in Tab.~\ref{tab:result_cityscapes} show a clear advantage over the uniform layout on all tested resolutions.
In all cases the rectilinear layouts outperformed their curvilinear counterparts, which might be because of their imitated adaptability in the image corners.
Fig.~\ref{fig:cityscapes_qualitative} also shows encouraging results for the detection of small objects in the image center, where objects are detected which the uniform layout misses.
\vspace{0.05cm}

\noindent\textbf{Multi-label classification\hspace{0.1cm}}
\begin{table}[]
    \centering
    \begin{tabular}{@{}c|c|c|c@{}}
         Resolution & Method &  Accuracy & $\theta$\\
         \hline
            \multirow{2}{*}{$8\times 8$} & ResNet18 & 86.60\%& - \\
     &ResNet18 + $\phi_{rect}$ & \textbf{87.19}\% & (0.32, 0.08) \\
     \hline
      \multirow{2}{*}{$16\times 16$} & ResNet18 & 88.39\%& - \\
         &ResNet18 + $\phi_{rect}$ & \textbf{88.62}\% & (0.15, 0.03)

    \end{tabular}
    \caption{Accuracy and learned pixel layouts for multi-label classification on the CelebA dataset \cite{liu2015faceattributes}.}
    \label{tab:res_celeba}
\end{table}
Finally, we evaluate our method on  multi-label classification of facial attributes on the CelebA dataset \cite{liu2015faceattributes}. 
Similarly to MNIST, the faces are in the image center, so that a non-standard pixel layout could be advantageous.
We fine-tune a ResNet18 pretrained on ImageNet on different resolutions over 15 epochs with rectilinear pixel layout.
We use a batch size of 64 and an Adam optimizer with a learning rate of $0.001$.

The results in Tab.~\ref{tab:res_celeba} show marginal improvements over the baselines.
As expected, the network learns a higher density in the image center. We conjecture that the joint prediction of 40 different attributes (spatially spread over each face) is the reason for the comparably small improvements.

\section{Conclusions}
\label{s:concl}
In this paper, we present the first approach of a parameterizable and differentiable pixel layout that can be jointly optimized with any downstream, task-specific network in an end-to-end fashion.
This approach allows going beyond fixed, regular pixel layouts that treat all regions of an image as equally important, yielding task specific layouts on which the networks perform better than on regular grids.

We provide the generic concept of an analytic, differentiable sensor layout parameterization that retains a regular topology and present two pixel layout parameterization functions, \ie, rectangular and curvilinear grid shapes.
Our drop-in module allows applying the learnable pixel layout to existing high-resolution imagery, and, thus, connecting our method to existing deep-learning pipelines.
We show that network predictions benefit from a learnable pixel layout for two different tasks, specifically multi-label classification and semantic segmentation.

\noindent\textbf{Limitations.} The proposed deformations are comparably simple and do not cover all task preferences. In addition, not all tasks have a sufficient spatial bias to make non-uniform sensor layouts advantageous,
 because such spatial biases (probabilistically) restrict the relative poses between objects of interest and the camera. Yet, we believe that sufficiently many application scenarios still naturally benefit from non-uniform layouts. 

{\small
\bibliographystyle{ieee_fullname}
\bibliography{egbib}
}

\end{document}